\PassOptionsToPackage{table}{xcolor}
\documentclass[]{antgroup}
\usepackage{antgroup}

\usepackage{latexsym}
\usepackage[T1]{fontenc}
\usepackage[utf8]{inputenc}
\usepackage{amsmath}
\usepackage{amssymb}
\usepackage{wrapfig}
\usepackage{array}
\usepackage{needspace}
\usepackage{xspace}
\usepackage[ruled,vlined]{algorithm2e}

\usepackage{amsmath,amsfonts,bm}

\def\eqref#1{equation~\ref{#1}}

\def\1{\bm{1}}

\DeclareMathAlphabet{\mathsfit}{\encodingdefault}{\sfdefault}{m}{sl}
\SetMathAlphabet{\mathsfit}{bold}{\encodingdefault}{\sfdefault}{bx}{n}

\DeclareMathOperator*{\argmax}{arg\,max}

\newcommand{\Mask}{[\mathrm{Mask}]}
\newcommand{\Del}{[\mathrm{Del}]}
\newcommand{\Ins}{[\mathrm{Ins}]}

\newcommand{\Vout}{\mathcal{V}_{\mathrm{out}}}
\newcommand{\Voc}{\mathcal{V}}

\newcommand{\Method}{LED}          
\newcommand{\Edit}{LLaDA-LED}      
\newcommand{\Base}{LLaDA2.1-Mini Base}

\DeclareMathOperator{\topop}{top}

\newcommand{\mtt}{M2T}
\newcommand{\ttt}{T2T}
\newcommand{\ROE}{ROE}

\definecolor{lightgray}{gray}{0.95}

\makeatletter
\newcommand{\tightenwrapfigurespace}{%
  \advance\WF@size -1.5\baselineskip
}
\makeatother
\title{Beyond Replacement: Levenshtein Editing for Diffusion Language Models}
\author{\centering
  Shuaidi Wang$^{1,2,\dagger}$, Zhuocheng Gong$^{2}$, Xuejie Liu$^{2}$,\\
  Jianguo Li$^{2,*}$, Yu Zhang$^{1,*}$
}
\affiliation{\centering
  $^1$ Southern University of Science and Technology\\
  $^2$ Inclusion AI, Ant Group
}

\begin{document}

\maketitle
\begingroup
\renewcommand{\thefootnote}{}
\footnotetext{\textsuperscript{*}Corresponding authors.\\
\hspace*{1.8em}\textsuperscript{\ensuremath{\dagger}}This work was done during Shuaidi Wang's internship at Ant Group.}
\endgroup

\begin{abstract}
Diffusion language models (DLMs) offer a flexible alternative to autoregressive models, enabling any-order infilling without specialized prompting. While most existing decoding methods treat decoded tokens as fixed, some recent models allow previously decoded tokens to be revised only by replacing them with alternative tokens, a paradigm we refer to as \emph{replacement-only editing} (\ROE{}).
Although \ROE{} can correct individual tokens, it remains inefficient for sequence shifts caused by missing or extra tokens and struggles when errors occur across many positions.
To address these limitations, we propose \Method{} (\textbf{L}evenshtein \textbf{E}diting for \textbf{D}iffusion Language Models), a post-training framework that extends \ROE{} with insertion and deletion, enabling Levenshtein-style refinement of generated sequences.
It uses two key training designs.
First, it reconstructs edit supervision on the fly during supervised fine-tuning using longest-common-subsequence alignment, enabling the model to learn insertion and deletion operations without manually annotated labels.
Second, it adopts multi-stage training, where the state from one stage becomes the input to the next, exposing the model to its own earlier errors and enabling it to learn how to correct them.
When applied to the 16B-parameter \Base{}, \Method{} raises the average score across twelve benchmarks from $54.86$ to $57.10$ under confidence-guided decoding while increasing tokens per forward from $6.14$ to $7.05$, whereas under confidence-free decoding it raises the average score from $27.52$ to $47.19$. We further show that, in this setting, \Method{} achieves higher pass@$k$ at large $k$ than thresholded decoding without editing, demonstrating improved generation diversity and revealing the diversity potential of DLMs without confidence guidance. 
\Method{} also improves robustness to large block sizes, raising the average score from $36.97$ to $51.34$ at block size $128$.
Together, these results suggest that \Method{} supports more robust self-correction in DLMs.
\end{abstract}

\section{Introduction}
\label{sec:intro}

Masked diffusion language models (DLMs) generate text over successive denoising rounds by repeatedly filling masked positions with predicted tokens, allowing multiple positions within a block to be decoded in parallel \citep{mdlm,shi2024md4,llada,dream,nie2025scaling,gong2025diffulama}.
We call this progressive mask-filling procedure mask-to-token (\mtt{}) generation.
However, standard \mtt{} generation cannot revisit an earlier prediction: once a token is decoded, later rounds use it as context but do not revise it.
Consequently, an error made early in decoding can affect subsequent predictions and remain in the final output.

Recent methods address this limitation by remasking or editing previously decoded tokens \citep{remdm,remedi,core,dreamon,hong2026wino,frkovic2026remasking,jeon2026mdmvgb}. 
Among these efforts, LLaDA2.1 is a prominent large-scale realization, with 16B Mini and 100B Flash variants \citep{llada21}.
Beyond \mtt{} generation, it introduces \emph{token-to-token} (\ttt{}) editing, which allows the model to replace an erroneous decoded token with a corrected token in a denoising round after its initial decoding.
We refer to this operation as \emph{replacement-only editing} (\ROE{}).
By enabling such corrections, \ROE{} supports more aggressive parallel drafting, improving the trade-off between generation quality and decoding speed.

\begin{figure}[!t]
  \centering
  \begin{minipage}[c]{0.43\linewidth}
    \centering
    \includegraphics[width=\linewidth]{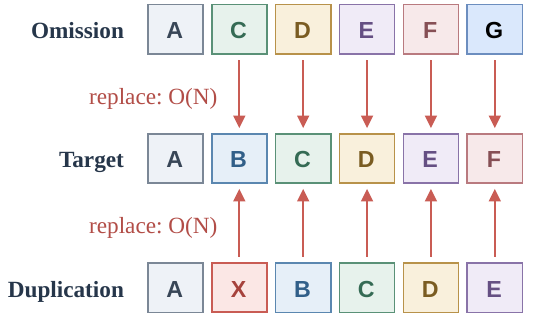}
  \end{minipage}\hfill
  \begin{minipage}[c]{0.5\linewidth}
    \centering
    \makebox[\linewidth][l]{\hspace*{-0.04\linewidth}\includegraphics[width=\linewidth]{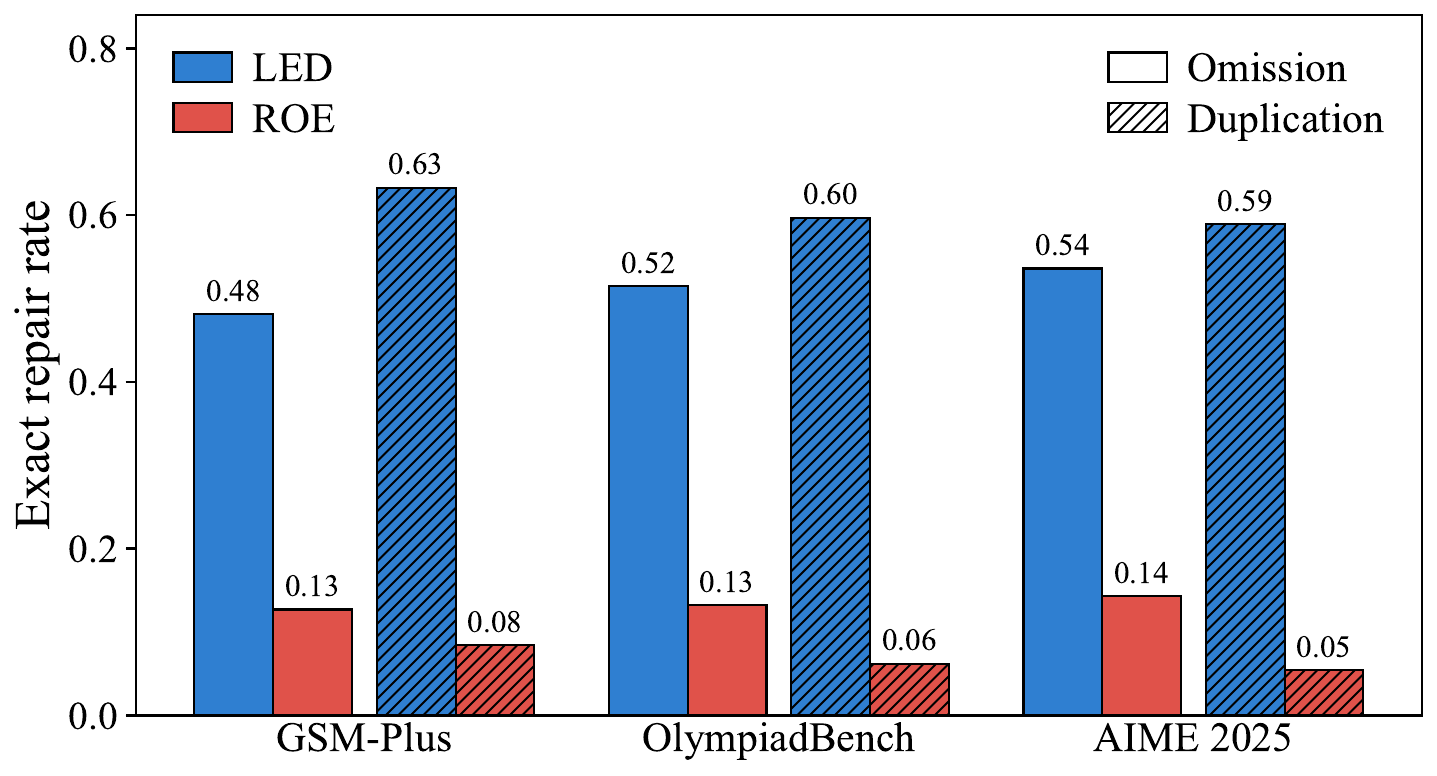}}
  \end{minipage}
  \caption{Sequence misalignment and repair. \emph{Left:} Schematic illustration of sequence misalignment. The center row is the target sequence, and the outer rows show fixed-length corrupted blocks after one token is omitted or duplicated, shifting the suffix and displacing a token at the block boundary. \emph{Right:} Exact repair rate under single-token sequence misalignment. The \ROE{} baseline achieves low repair rates, whereas \Method{} yields substantially higher rates. Experimental details are provided in Appendix~\ref{app:misalignment}.}
  \label{fig:misalign}
\end{figure}

Despite these benefits, \ROE{} exhibits two failure modes that limit its editing capability.
First, \ROE{} cannot efficiently correct sequence misalignment, which arises when token omission or duplication shifts the remaining suffix relative to the target sequence.
This inefficiency stems from the fact that \ROE{} can only replace tokens at their current positions, so realigning a suffix of length $N$ may require replacing all $N$ tokens, as illustrated in the left panel of Figure~\ref{fig:misalign}.
This $\mathcal{O}(N)$ replacement cost makes reliable correction difficult and can consume additional denoising rounds, limiting both generation quality and decoding speed.
The right panel of Figure~\ref{fig:misalign} empirically illustrates this difficulty.
Second, \ROE{} remains ineffective when errors occupy a large portion of the sequence.
To expose this failure mode, we decode every masked position in a block in a single forward pass without confidence filtering, after which the decoder must rely solely on \ROE{} to repair the resulting draft.
The resulting performance drops substantially (Section~\ref{sec:exp-confidence-free}), indicating that \ROE{} cannot reliably recover from widespread errors.

To address these limitations, we propose \textbf{L}evenshtein \textbf{E}diting for \textbf{D}iffusion Language Models (\Method{}), a post-training framework that extends \ROE{} with insertion and deletion, enabling Levenshtein-style refinement of generated sequences.
\Method{} models insertion and deletion by introducing a special output token for each operation.
The training procedure in \Method{} has two key designs. 
First, \Method{} reconstructs edit supervision online via longest-common-subsequence (LCS) alignment, enabling the new edit actions to be learned from standard supervised fine-tuning (SFT) data.
Second, \Method{} constructs multi-stage training trajectories in which later stages receive states partly formed by model predictions from earlier stages.
This exposes the model to its own intermediate errors and makes the training states more similar to those encountered during inference.

We post-train \Base{} with \Method{} for 3.7k H200 GPU-hours to obtain \Edit{}.
Under confidence-guided decoding, which decodes only masked positions whose confidence exceeds a predefined threshold, \Edit{} raises the average score from $54.86$ to $57.10$ across twelve generation benchmarks while improving decoding parallelism (Section~\ref{sec:exp-main}).
Under confidence-free decoding, every masked position is decoded in a single forward pass without confidence filtering, so the model must rely solely on its own editing operations in later rounds to correct errors in the resulting draft. In this setting, \Edit{} substantially raises the average score from $27.52$ to $47.19$ (Section~\ref{sec:exp-confidence-free}).
This substantial gain indicates that \Method{} markedly strengthens DLM self-correction when drafts contain widespread errors.

Beyond strengthening self-correction, combining \Method{} with confidence-free
decoding also addresses a limitation of confidence-guided decoding. As the most
commonly used decoding strategy for DLMs, confidence-guided decoding commits
the most confident positions first and defers uncertain ones. Prior work
observes that this preference postpones high-entropy positions that may be
crucial for reaching a correct answer
\citep{ni2026flexibilitytrap,olausson2026temperatures}. We call this tendency
to defer high-entropy positions the \emph{low-entropy bias}. Simply removing
confidence filtering is insufficient, because errors introduced at uncertain
positions can propagate through subsequent rounds and sharply degrade output
quality. By repairing these errors, \Method{} mitigates the quality loss
associated with confidence-free decoding. Consistent with this account,
\Edit{} with confidence-free decoding achieves higher pass@$k$ at large $k$
than thresholded decoding without editing
(Section~\ref{sec:exp-passk-decoding}). This result
shows that pairing \Method{} with confidence-free decoding can turn
higher-entropy exploration into broader solution coverage, suggesting a viable
path toward confidence-free decoding.
Beyond these results, we evaluate the robustness of \Method{} to larger block
sizes (Section~\ref{sec:exp-block}) and offer an entropy-based interpretation
of its editing behavior. The main contributions of this work are three-fold.
\begin{itemize}
\item We identify two limitations of replacement-only editing in DLMs: difficulty correcting sequence misalignments and limited ability to recover from widespread errors in intermediate states.

\item We propose \Method{}, a post-training framework that equips DLMs with insertion and deletion capabilities through online reconstruction of LCS-based edit supervision and inference-aligned multi-stage training.

\item We validate \Method{} across twelve benchmarks and identify three benefits. With confidence guidance, it raises the average score while increasing tokens per forward. Under confidence-free decoding, it improves pass@$k$ at large sampling budgets, indicating broader solution coverage. It also improves robustness to larger block sizes.
\end{itemize}

\section{Related Work}
\label{sec:related}

\textbf{Diffusion language models and decoding.}
Masked DLMs have been scaled to large language models \citep{mdlm,shi2024md4,llada,dream}.
Recent work studies confidence-guided, learned, and block-wise decoding policies to balance generation quality and parallelism \citep{fastdllm2025,learning-unmasking,sparsedllm,hu2026rcd}.

\textbf{Editing and refinement for sequence generation.}
Existing DLM revision methods remask or replace earlier predictions, whereas edit-based and flexible-length models use insertion and deletion \citep{remdm,remedi,core,llada21,levenshtein-transformer,havasi2025edit,dreamon,kim2025flexmdm,ding2026did}.
Among other edit-based methods, ME-DLM applies sequence edits after completing a diffusion draft, with an additional next-token head and a three-stage training pipeline \citep{ren2026editbased}.
\Method{} instead learns token prediction, deletion, and insertion in one output vocabulary along the denoising trajectory, allowing alignment correction before block generation is complete while retaining the fixed-block interface of the base DLM.
This editing method has also been applied to the LLaDA2.2 model series~\citep{llada22}.
Further comparisons with other edit-based methods appear in Appendix~\ref{app:related-comparison}.

\section{Background}
\label{sec:background}
We only consider the block diffusion setting in this paper, in which a DLM generates text autoregressively in fixed-size blocks while predicting the tokens within each block in parallel over iterative denoising rounds \citep{mdlm,shi2024md4,arriola2025block}. Let $x=(x_1,\ldots,x_B)\in\Voc^B$ denote a clean token block of length $B$, where $\Voc$ is the model's token vocabulary and $\Mask\notin\Voc$ is a special mask token. During training, the forward process progressively masks the clean block by replacing selected tokens with $\Mask$. Given a masking rate $t\in[0,1]$, we independently sample $m_i^t\sim\mathrm{Bernoulli}(t)$ and construct the partially masked block $\tilde{x}^t$ as
\begin{equation}
\tilde{x}_i^t =
\begin{cases}
\Mask, & m_i^t=1,\\
x_i, & m_i^t=0,
\end{cases}
\qquad i=1,\ldots,B.
\end{equation}
The model is then trained to recover the original tokens at the masked positions conditioned on the partially masked block $\tilde{x}^t$.
At decoding round $r$, let $s^r=(s_1^r,\ldots,s_B^r)\in(\Voc\cup\{\Mask\})^B$ denote the current block state. We define the sets of masked and non-masked positions as
\begin{equation}
\mathcal{M}_r=\{i\in\{1,\ldots,B\}:s_i^r=\Mask\},
\qquad
\mathcal{P}_r=\{i\in\{1,\ldots,B\}:s_i^r\in\Voc\}.
\end{equation}

When discussing decoding strategies, a common choice is confidence-based decoding, which commits masked positions whose predicted tokens are sufficiently reliable after each forward pass. Let $\hat{v}_i^r$ denote the token predicted at masked position $i$ in round $r$, with confidence $p_\theta(\hat{v}_i^r\mid c,s^r,i)$. Given a confidence threshold $\tau$, the decoder commits the masked positions whose confidence exceeds $\tau$, i.e., $\{i\in\mathcal{M}_r : p_\theta(\hat{v}_i^r\mid c,s^r,i)\ge\tau\}$. If no position satisfies the threshold, the highest-confidence masked position can be selected to ensure progress.
When $\tau=0$, all masked positions are decoded in one forward pass without confidence-based selection, which we refer to as confidence-free decoding in this paper.

\section{Methodology}
\label{sec:method}

This section presents \Method{}'s framework for Levenshtein-style editing in DLMs. It combines token replacement, insertion, and deletion, the three basic operations of Levenshtein editing. We first describe how the model represents these operations as position-aligned actions (Section~\ref{sec:method-transition}). We then describe how the model learns to predict these actions within a standard SFT pipeline, with LCS alignment providing the edit supervision (Section~\ref{sec:method-lcs}). Finally, we introduce multi-stage training to narrow the training--inference gap in learning insertion and deletion (Section~\ref{sec:method-multistage}).

\subsection{Modeling Insertion and Deletion}
\label{sec:method-transition}

Insertion and deletion are enabled by extending the output vocabulary with two special tokens, $\Del$ and $\Ins$, so that $\Vout=\Voc\cup\{\Del,\Ins\}$. Both tokens are excluded from the input vocabulary. Given the input $s^r$ at denoising round $r$, the model predicts one token from $\Vout$ at each position. At each selected position $i$, a prediction of $\Del$ removes $s_i^r$, while a prediction of $\Ins$ inserts a fresh $\Mask$ immediately before it. As these edits can change the sequence length, the resulting sequence is truncated to $B$ tokens or right-padded with $\Mask$ tokens before serving as the input for round $r+1$.

\subsection{LCS-Based Edit Supervision}
\label{sec:method-lcs}

Standard SFT data does not provide direct supervision signals for insertion and deletion. To train the special tokens $\Del$ and $\Ins$, we reconstruct position-wise edit targets online during the SFT stage. Given a current source state $s^r$ and its corresponding clean target $y=(y_1,\ldots,y_B)\in\Voc^B$, the reconstruction assigns each selected source position $i$ an edit target $e_i^r\in\Voc\cup\{\Del,\Ins\}$. The reconstruction therefore enables insertion and deletion to be learned within the standard SFT pipeline without manually annotated edit trajectories.

\begin{figure}[!t]
\centering
\includegraphics[width=0.85\linewidth]{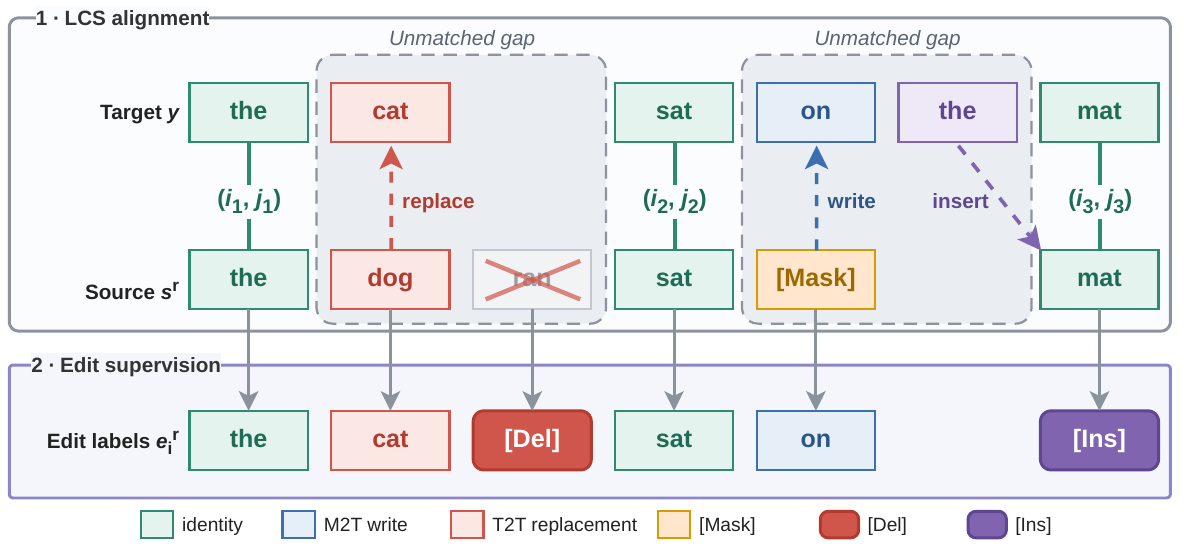}
\caption{LCS-based edit-target construction. Green links denote matched anchor pairs $(i_k,j_k)$. Within each unmatched gap, surplus source positions receive $\Del$, and at most one surplus target token is handled by assigning $\Ins$ to the matched source anchor immediately to its right.}
\label{fig:lcs}
\end{figure}

We construct the edit targets by aligning $s^r$ with $y$ through longest-common-subsequence (LCS) matching. Let $\pi^\star=\{(i_k,j_k)\}_{k=1}^{K}$ be a maximum-cardinality monotone matching of source and target indices, where $i_k,j_k\in\{1,\ldots,B\}$ and $s_{i_k}^r=y_{j_k}\in\Voc$ for each matched pair. We leave the supervision signals at these matched positions unchanged, thereby preserving as much correct information in the source state as possible and concentrating edit supervision on the unmatched regions. Figure~\ref{fig:lcs} illustrates the matched anchors and the resulting edit targets.

The matched anchors partition the unmatched source and target positions into $K+1$ ordered gaps. For convenience, let $i_0=j_0=0$ and $i_{K+1}=j_{K+1}=B+1$, and define the source and target positions in gap $k$ as $I_k=\{i_k+1,\ldots,i_{k+1}-1\}$ and $J_k=\{j_k+1,\ldots,j_{k+1}-1\}$, respectively, for $k=0,\ldots,K$. As described above, for each matched pair $(i_k,j_k)$, we keep its supervision unchanged by setting $e_{i_k}^r=y_{j_k}$, unless the source anchor is used to supervise an insertion as described below. Within each gap, let $\ell_k=\min\{|I_k|,|J_k|\}$. We first pair the first $\ell_k$ source and target positions in order and assign $e_{I_k[p]}^r=y_{J_k[p]}$ for $p=1,\ldots,\ell_k$, where $I_k[p]$ and $J_k[p]$ denote the $p$-th positions in the corresponding ordered gaps. If $|I_k|>\ell_k$, the remaining source positions correspond to surplus tokens and are assigned deletion targets, i.e., $e_{I_k[p]}^r=\Del$ for $p=\ell_k+1,\ldots,|I_k|$. If $|J_k|>\ell_k$, the remaining target positions require insertion; we therefore assign $e_{i_{k+1}}^r=\Ins$ to the matched source anchor immediately to the right, provided that $k<K$. Each gap produces at most one $\Ins$ target per round.

The reconstruction is performed independently for every block. Exact LCS alignment costs $\mathcal{O}(B^2)$ time for a block of fixed length $B$, giving a total cost of $\mathcal{O}(N_{\mathrm{blk}}B^2)$ where $N_{\mathrm{blk}}$ is the number of blocks. The online cost therefore grows linearly with the number of blocks when $B$ is fixed. This makes it practical to integrate insertion and deletion supervision into standard SFT. Appendix~\ref{app:algs} provides the complete gap-level rules and Algorithm~\ref{alg:lcs}.

\subsection{Edit-Augmented Inference}
\label{sec:method-infer}

Our inference procedure follows the confidence-based decoding strategy in Section~\ref{sec:background}. At round $r$, the decoder selects masked positions according to threshold $\tau$, using the highest-confidence masked position as a fallback when no position meets the threshold. We denote this set by $\mathcal{C}_r(\tau)$. However, $\mathcal{C}_r(\tau)$ only determines which masked positions are decoded and does not allow already generated tokens to be revised. We therefore introduce an edit threshold $\nu$ for selecting non-masked positions. For each $i\in\mathcal{P}_r$, let $\hat{v}_i^r$ denote the model's predicted output, with confidence $\eta_i^r=p_\theta(\hat{v}_i^r\mid c,s^r,i)$. We select a non-masked position only when the model proposes an output different from its current token with confidence at least $\nu$, yielding $\mathcal{E}_r(\nu)=\{i\in\mathcal{P}_r:\hat{v}_i^r\neq s_i^r,\ \eta_i^r\geq\nu\}$. The positions updated in round $r$ are therefore $\mathcal{Q}_r=\mathcal{C}_r(\tau)\cup\mathcal{E}_r(\nu)$.

The decoder updates only the positions in $\mathcal{Q}_r$ and retains all remaining positions. Predictions of $\Del$ and $\Ins$ are processed as defined in Section~\ref{sec:method-transition}.

\subsection{Inference-Aligned Multi-Stage Training}
\label{sec:method-multistage}

During inference, the input to each round depends on the model's predictions in preceding rounds. In contrast, standard training constructs intermediate states through random noising rather than the model's own decoding process, leaving the model unexposed to the states it actually encounters at inference time. To narrow this training--inference gap, we construct multi-stage training trajectories that mimic the inference process.

\begin{figure}[!t]
\centering
\includegraphics[width=0.92\linewidth]{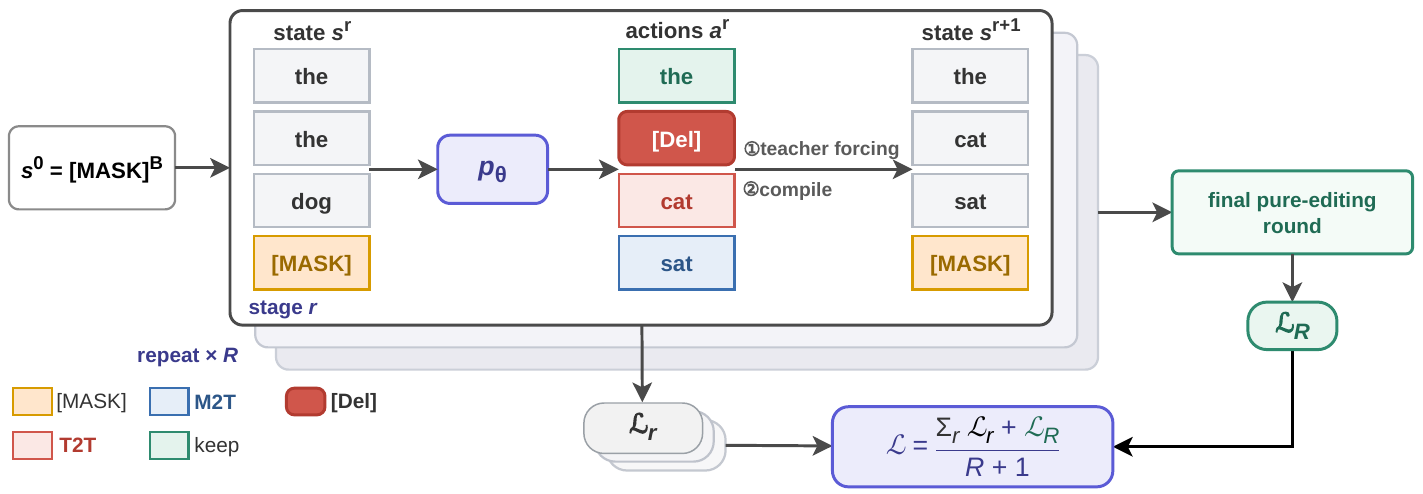}
\caption{Inference-aligned multi-stage training. Each stage supervises selected masks and visible positions. A final editing-only stage covers the visible positions after the last mask-update stage.}
\label{fig:multistage}
\end{figure}

Specifically, we perform multiple forward passes on each training example. In the input to the first stage, all response tokens are masked, whereas in the input to the final stage, all response tokens have been decoded. The output of each stage influences the input to the subsequent stage. Figure~\ref{fig:multistage} illustrates this multi-stage construction, and Appendix~\ref{app:algs} provides the complete procedure.

In practice, using the model output from one stage directly as the input to the next makes training challenging, especially early in training when its predictions are still unreliable. We therefore use mixed teacher forcing to construct stable trajectories while still exposing the model to its own predictions. At stage $r$, LCS provides labels $e_i^r\in\Vout$ for $\mathcal{Q}_r=\mathcal{U}_r\cup\mathcal{P}_r$. For each selected masked position $i\in\mathcal{U}_r$, we randomly mix its LCS-derived label and the model prediction when forming the next-stage input.
For non-masked positions, we set $\tilde{v}_i^r=e_i^r$. We apply the selected predictions to form $s^{r+1}$, which serves as the input to the next stage. Using LCS-derived labels stabilizes the early stages, whereas using model predictions exposes later stages to model-generated inputs.

At each stage, we train the model to predict the LCS-derived target $e_i^r$ at every selected position $i\in\mathcal{Q}_r$, conditioned on the context $c$ and current response state $s^r$. We use the average negative log-likelihood over these selected positions:
\begin{equation}
\mathcal{L}_r =
-\frac{1}{|\mathcal{Q}_r|}
\sum_{i\in\mathcal{Q}_r}
\log p_\theta(v=e_i^r\mid c,s^r,i).
\label{eq:stage_r_loss}
\end{equation}
Thus, each stage encourages the model to recover the edit targets specified by LCS from its current intermediate state. Normalizing by $|\mathcal{Q}_r|$ gives each stage equal weight regardless of its number of selected positions.

\section{Experiments and Analysis}
\label{sec:exp}

\subsection{Setup}
\label{sec:exp-setup}
We post-train \Base{} \citep{llada21} on 745k SFT examples. The post-training run for \Edit{} consumes 3.7k H200 GPU-hours. The replacement-only baseline (w/o \Method{}) and \Edit{} (w/ \Method{}) use the same base model and data, differing only in whether \Method{} is included in post-training. We set $R=4$ and teacher forcing ratio to $0.5$. Unless stated otherwise, evaluation uses a block size of 32 and a maximum generation length of 32,768. Appendix~\ref{app:hyper} provides the remaining hyperparameters. We evaluate twelve benchmarks spanning seven capability categories and report per-benchmark scores, their unweighted average, and tokens per forward (TPF), where higher TPF indicates greater decoding parallelism. Appendix~\ref{app:bench} provides benchmark and scoring details.

\subsection{Main Results under Confidence-Guided Decoding}
\label{sec:exp-main}

\begin{table}[t]
\caption{Benchmark results under confidence-guided decoding with the balanced threshold $\tau=0.5$. We compare the baseline (w/o \Method{}) with our method (w/ \Method{}). Higher score and TPF are better. Bold indicates the higher value within each metric pair.}
\label{tab:main}
\centering
\small
\setlength{\tabcolsep}{5pt}
\renewcommand{\arraystretch}{1.15}
\begin{tabular}{l l c c c c}
\toprule
\multirow{2}{*}{Category} & \multirow{2}{*}{Benchmark} & \multicolumn{2}{c}{Score $\uparrow$} & \multicolumn{2}{c}{TPF $\uparrow$} \\
\cmidrule(lr){3-4}\cmidrule(lr){5-6}
& & w/o \Method{} & w/ \Method{} & w/o \Method{} & w/ \Method{} \\
\midrule
\multirow{2}{*}{Knowledge} & MMLU-Pro   & 53.70 & \textbf{55.33} & 4.53 & \textbf{5.60} \\
                           & GPQA-D     & 39.65 & \textbf{40.91} & 4.14 & \textbf{5.74} \\
\midrule
Reasoning & KOR-Bench  & 32.88 & \textbf{44.00} & 6.80 & \textbf{8.94} \\
\midrule
\multirow{3}{*}{Math} & GSM+       & \textbf{83.57} & 83.06 & 4.13 & \textbf{4.67} \\
                      & Olympiad   & \textbf{55.85} & 54.11 & 5.19 & \textbf{5.32} \\
                      & AIME 25    & \textbf{30.00} & 24.01 & 4.52 & \textbf{5.02} \\
\midrule
\multirow{3}{*}{Coding} & HumanEval+ & 76.22 & \textbf{79.19} & \textbf{8.01} & 7.49 \\
                        & MBPP+      & 65.08 & \textbf{70.90} & 7.75 & \textbf{8.78} \\
                        & LCB v6     & 13.88 & \textbf{24.34} & 8.32 & \textbf{10.55} \\
\midrule
Function Calling & BFCL v3    & \textbf{65.39} & 64.85 & \textbf{9.15} & 6.80 \\
\midrule
Inst. Following & IFEval     & \textbf{69.69} & 68.39 & 4.24 & \textbf{6.03} \\
\midrule
Chinese & C-Eval     & 72.36 & \textbf{76.15} & 6.97 & \textbf{9.69} \\
\midrule
\multicolumn{2}{l}{Avg} & 54.86 & \textbf{57.10} & 6.14 & \textbf{7.05} \\
\bottomrule
\end{tabular}
\end{table}

Table~\ref{tab:main} reports the main results under confidence-guided decoding with the default balanced threshold $\tau=0.5$.
The baseline and \Edit{} share the same base model and training data. They differ only in whether \Method{} is applied during post-training.

At this setting, \Method{} raises the unweighted average score across the twelve benchmarks from $54.86$ to $57.10$, a gain of $2.24$ points, while increasing TPF from $6.14$ to $7.05$. The score improvement is task-dependent, with gains on seven of the twelve benchmarks, including the largest increases on KOR-Bench and LiveCodeBench v6. Together, these results indicate that the average score gain does not come at the expense of decoding parallelism. The joint improvement is consistent with the design of \Method{}, in which insertion and deletion can repair local sequence misalignments without replacing an entire shifted suffix.

\subsection{Confidence-Free Decoding}
\label{sec:exp-confidence-free}

At $\tau=0$, every masked position in a block is decoded in a single forward pass without confidence filtering, increasing within-block parallelism. However, decoding uncertain positions simultaneously can yield intermediate states with widespread errors, requiring subsequent rounds to correct them through editing operations. 

\begin{figure}[!t]
    \centering
    \includegraphics[width=0.96\linewidth]{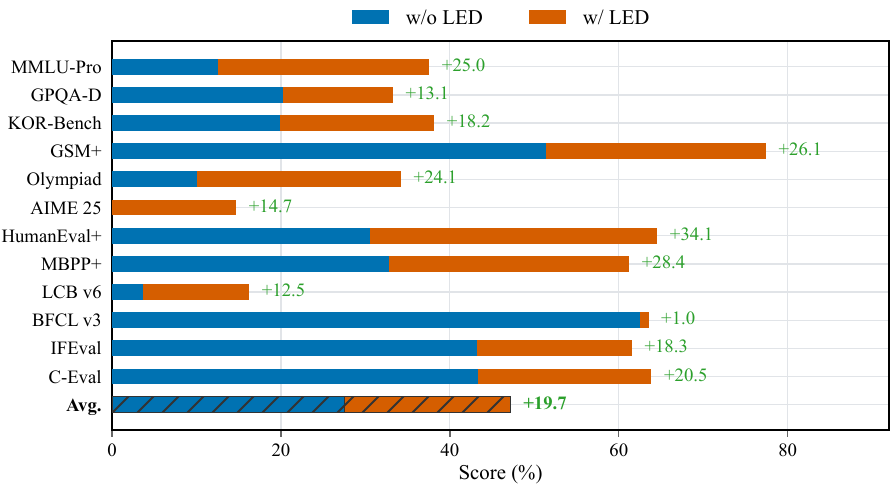}
    \caption{Per-benchmark scores under confidence-free decoding with threshold $\tau=0$. Diagonal hatching highlights the average row.}
    \label{fig:confidence-free-score}
\end{figure}

As shown in Figure~\ref{fig:confidence-free-score}, the replacement-only baseline attains an average score of $27.52$, revealing the difficulty of recovering from widespread errors with replacement-only editing. Applying \Method{} improves the score on all twelve benchmarks. The average score increases from $27.52$ to $47.19$, a gain of $19.67$ points. Per-benchmark gains range from $+0.98$ on BFCL v3 to $+34.07$ on HumanEval+. The largest improvements occur on HumanEval+, MBPP+, GSM+, and MMLU-Pro. Improvements across all twelve benchmarks span knowledge, reasoning, math, coding, function-calling, instruction-following, and Chinese evaluations, suggesting that \Method{} helps recover from drafts containing widespread errors.

This result is consistent with the roles of insertion and deletion in repairing length and alignment errors, together with inference-aligned training that exposes the model to model-induced intermediate states. Moreover, these results support the claim that \Method{} makes confidence-free decoding a more viable regime and motivate the solution-coverage analysis in Section~\ref{sec:exp-passk-decoding}.

\subsection{Confidence-Free Decoding Broadens Solution Coverage}
\label{sec:exp-passk-decoding}

\begin{figure}[htbp]
\centering
\includegraphics[width=\linewidth]{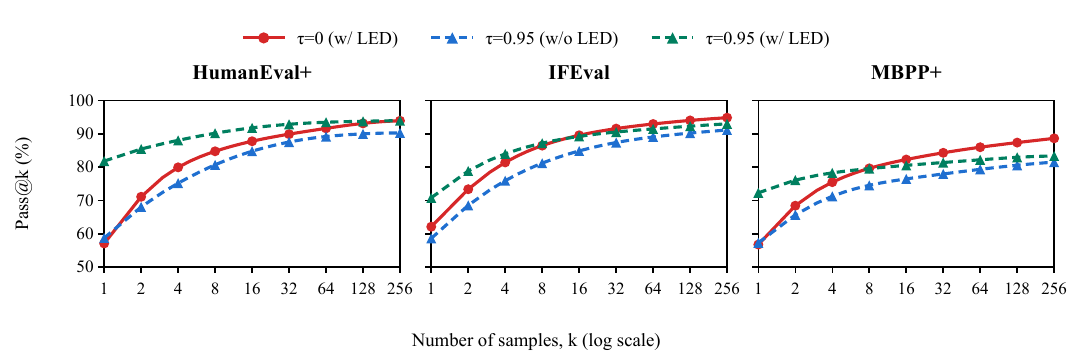}
\caption{Pass@$k$ under confidence-free and confidence-guided decoding on HumanEval+, IFEval, and MBPP+, using 256 samples per problem. Red denotes confidence-free decoding with \Method{} ($\tau=0$); blue and green denote confidence-guided decoding without editing and with \Method{}, respectively ($\tau=0.95$).}
\label{fig:passk-decoding-regimes}
\end{figure}

Confidence-guided decoding updates only masked positions whose predicted-token confidence exceeds the threshold $\tau$. At a high threshold such as $\tau=0.95$, it favors locally concentrated predictions and defers uncertain positions. This preference can cause independently sampled decoding trajectories to concentrate on similar regions of the solution space~\citep{ni2026flexibilitytrap,olausson2026temperatures}. Confidence-free decoding removes this selection rule and resolves all masked positions in each round, thereby admitting decisions that would otherwise be postponed. However, simultaneously committing uncertain predictions can create noisy and locally misaligned drafts. \Method{} addresses this difficulty by enabling later rounds to revise token content and repair local sequence errors, allowing the decoder to retain potentially useful alternatives while correcting the errors introduced during parallel decoding. The remaining question is whether this repair capability converts the additional choices made under confidence-free decoding into useful solution coverage.

We evaluate this question using Pass@$k$, the probability that at least one of $k$ sampled outputs is successful. Figure~\ref{fig:passk-decoding-regimes} compares confidence-free decoding with \Method{} at $\tau=0$ against confidence-guided decoding at $\tau=0.95$, both without editing and with \Method{}. These three configurations evaluate the contribution of editing under confidence-guided decoding and the effect of removing confidence-based position selection when editing is available. Each configuration generates 256 samples per problem.

At $\tau=0.95$, adding \Method{} improves Pass@$256$ over confidence-guided decoding without editing on all three benchmarks. Holding \Method{} fixed, confidence-guided decoding achieves higher Pass@$1$, whereas confidence-free decoding catches up as the sampling budget increases. At $k=256$, confidence-free decoding matches confidence-guided decoding with \Method{} on HumanEval+ and exceeds it by $1.85$ points on IFEval and $5.29$ points on MBPP+, for an average gain of $2.38$ points. These results show that editing supports both decoding regimes and that confidence-free decoding can broaden solution coverage when multiple samples are available.

\subsection{Entropy-Based Analysis of \Method{}'s Editing Behavior}
\label{sec:exp-edit-entropy}
\begin{wrapfigure}[17]{r}{0.46\textwidth}
\centering
\includegraphics[width=\linewidth]{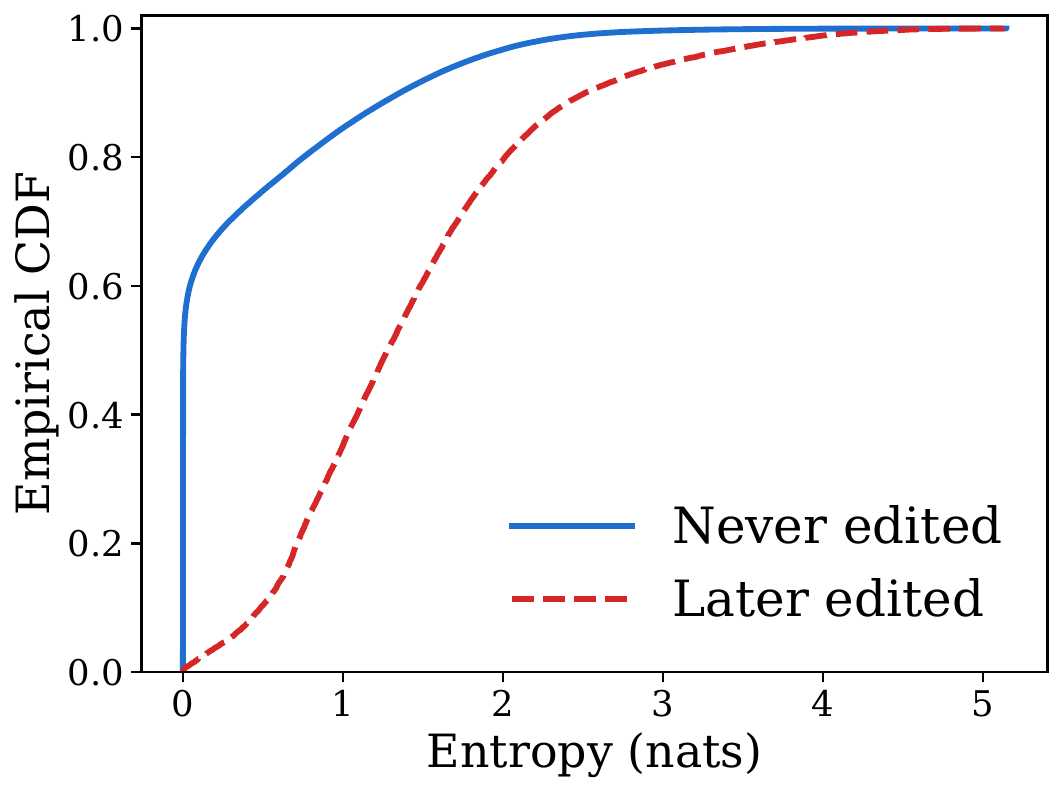}
\caption{Initial predictive entropy CDFs for MBPP+ tokens, grouped by later-edit status.}
\label{fig:entropy-ecdf}
\end{wrapfigure}

We examine whether \Method{} later edits tokens that were uncertain when first decoded by \mtt{}. Across 3,024 MBPP+ predictions, we compute each token's initial entropy over $\Vout$ as $H_i^r = -\sum_{v\in\Vout} p_\theta(v\mid c,s^r,i)\log p_\theta(v\mid c,s^r,i)$.

We trace each decoded token through subsequent operations and mark it as edited when it is modified in a later round. This definition accommodates insertions and deletions that change token positions.

Figure~\ref{fig:entropy-ecdf} compares the empirical cumulative distribution functions (CDFs) of initial predictive entropy for later-edited and never-edited tokens. The distribution for later-edited tokens is shifted toward higher entropy values, indicating that tokens selected for subsequent editing tend to have more uncertain initial output distributions. This comparison describes an association between initial predictive uncertainty and later editing.

\subsection{Robustness to Large Block Sizes}
\label{sec:exp-block}
\newsavebox{\blocktablebox}
\setlength{\tabcolsep}{4pt}
\sbox{\blocktablebox}{%
\begin{tabular}{llcccc}
\toprule
\multirow{2}{*}{Metric} & \multirow{2}{*}{Method} & \multicolumn{4}{c}{Block size} \\
\cmidrule(lr){3-6}
& & 32 & 64 & 128 & 256 \\
\midrule
\multirow{3}{*}{Score}
& w/ \Method{}  & 57.10 & 56.28 & 51.34 & 13.19 \\
& w/o \Method{} & 54.86 & 53.86 & 36.97 & 6.45 \\
& \quad \cellcolor{lightgray}\textit{diff} 
& \cellcolor{lightgray}\textbf{+2.24} 
& \cellcolor{lightgray}\textbf{+2.42} 
& \cellcolor{lightgray}\textbf{+14.37} 
& \cellcolor{lightgray}\textbf{+6.74} \\
\midrule
\multirow{3}{*}{TPF} 
& w/ \Method{}  & 7.05 & 7.85 & 8.63 & 8.09 \\
& w/o \Method{} & 6.14 & 6.17 & 9.57 & 7.65 \\
& \quad \cellcolor{lightgray}\textit{diff} 
& \cellcolor{lightgray}\textbf{+0.91} 
& \cellcolor{lightgray}\textbf{+1.68} 
& \cellcolor{lightgray}-0.94 
& \cellcolor{lightgray}\textbf{+0.44} \\
\bottomrule
\end{tabular}%
}

\begin{table}[!htbp]
\centering
\caption{Comparison between \Edit{} and the baseline across different block sizes.}
\label{tab:block}
\usebox{\blocktablebox}
\end{table}

Larger blocks create longer within-block denoising chains, over which early local errors can affect subsequent predictions. We therefore evaluate robustness to test-time block-size expansion: both compared models are trained with a block size of $32$ and evaluated at larger block sizes without retraining. Table~\ref{tab:block} shows modest gains from \Method{} at block sizes $32$ and $64$, but a pronounced divergence at $128$: \Edit{} attains an average score of $51.34$, compared with $36.97$ for the baseline. Although the baseline achieves a slightly higher TPF at this block size ($9.57$ versus $8.63$), this additional parallelism coincides with a $14.37$-point score deficit. At block size $256$, both methods degrade substantially, yet \Edit{} retains an average score of $13.19$ versus $6.45$ for the baseline. The widening score gap under test-time block-size expansion is consistent with insertion and deletion alleviating the accumulation of local errors.

\subsection{Grid Search over the Mask and Edit Confidence Thresholds}
\label{sec:exp-grid}

We sweep the mask confidence threshold $\tau\in\{0, 0.2, 0.4, 0.6, 0.8, 1.0\}$ and the edit confidence threshold $\nu\in\{0, 0.2, 0.4\}$, with the block size fixed at $32$. As defined in Sections~\ref{sec:background} and~\ref{sec:method-infer}, $\tau$ determines which masked positions are decoded in each round, whereas $\nu$ determines whether a non-masked position whose predicted output differs from its current token is selected for editing. Figure~\ref{fig:grid} in Appendix~\ref{app:training_ablation} shows a clear score--parallelism trade-off along the $\tau$ axis. Lower values of $\tau$ admit more masked positions per forward pass and yield higher TPF, but produce lower average scores. Increasing $\tau$ improves the average score while reducing TPF. In contrast, varying $\nu$ has a smaller effect on the aggregate metrics, with the largest score variation occurring at $\tau=0$, where all masked positions are decoded without confidence filtering. For the default evaluation configuration, we use $\tau=0.5$ as a balanced setting and set $\nu=0.0$, which accepts every proposed non-identity edit.

\section{Conclusion and Future Work}
\label{sec:conclusion}

We proposed \Method{}, a post-training framework that augments replacement-only editing with deletion and insertion, using online LCS alignment and inference-aligned trajectories to teach a DLM to repair length and alignment errors. Across confidence-free decoding and large-block settings, \Method{} substantially improves the average score, and controlled corruption probes support the role of length-changing edits in recovering from misaligned intermediate states. These results also suggest that confidence-free decoding merits further study: with \Method{}, it achieves higher Pass@$k$ than thresholded decoding without editing at large sampling budgets on the three evaluated benchmarks, indicating broader solution coverage. The current framework nevertheless constructs edit supervision locally at each denoising stage, so it does not explicitly optimize the quality or efficiency of the complete edit trajectory. Future work could therefore redesign insertion and deletion construction around path-level objectives and use trajectory-level learning, including reinforcement learning \citep{zhao2025d1,lightningrl} or on-policy distillation (OPD) \citep{agarwal2024opd,zhu2026copsd}, to learn when and how to coordinate these edits across denoising rounds. Such approaches may further improve the robustness and efficiency of confidence-free DLM decoding.

\section*{Acknowledgments}
This work was supported by the Ant Group Research Fund.
This work was partially supported by National Natural Science Foundation of China under Grant no. 62136005 and Shenzhen fundamental research program JCYJ20250604144724032.

\bibliography{iclr2027_conference}
\bibliographystyle{antgroup}

\appendix
\section{Implementation Details}
\label{app:impl}
This section provides supplementary experimental details and pseudocode for the procedures described in Section~\ref{sec:method}.

\subsection{Main Hyperparameters}
\label{app:hyper}
Table~\ref{tab:hyper} summarizes the configuration of our main experiments. These settings are shared by all compared systems unless a specific study states otherwise. All systems are post-trained from the same \Base{} on an identical SFT corpus and use the same optimization schedule. Note that an editing threshold of $0.0$ means that the token selected by self-correction is always accepted.
\begin{table}[h]
\caption{Main training and evaluation hyperparameters.}
\label{tab:hyper}
\begin{center}
\small
\renewcommand{\arraystretch}{1.15}
\begin{tabular}{@{}l l l@{}}
\toprule
Group & Hyperparameter & Value \\
\midrule
\multirow{2}{*}{Data}
  & Number of training samples & $745k$ \\
  & Total tokens per epoch     & $1.6$B \\
\midrule
\multirow{9}{*}{Training}
  & Training epochs            & $10$ \\
  & Maximum sequence length    & $16{,}384$ \\
  & Batch size                 & $256$ \\
  & Learning rate              & $1.0\times10^{-4}$ \\
  & Learning-rate schedule     & Cosine decay \\
  & Minimum learning rate      & $1.0\times10^{-6}$ \\
  & Maximum denoising stages   & $4$ \\
  & \mtt{} confidence threshold & $0.5$ \\
  & Teacher-forcing ratio      & $0.5$ \\
\midrule
\multirow{6}{*}{Evaluation}
  & Maximum new tokens         & $32{,}768$ \\
  & Temperature & $0.0$ \\
  & \mtt{} threshold           & $0.5$\\
  & \ttt{} threshold           & $0.0$\\
  & Maximum post-mask editing rounds & $16$\\
  & Block size                 & $32$\\
\bottomrule
\end{tabular}
\end{center}
\end{table}

\subsection{Misalignment-Repair Configuration}
\label{app:misalignment}
The controlled probe in Figure~\ref{fig:misalign} compares \Method{} with an \ROE{} baseline trained under the same one-forward schedule. We evaluate on three math reasoning datasets---GSM-Plus, OlympiadBench, and AIME 2025---using clean blocks that the baseline previously generated correctly. We use each clean block as the target sequence, which biases the comparison toward the baseline because the repair targets lie within its output distribution. At a seeded position, we corrupt each target block by omitting one token or inserting a random token. Either operation shifts the remaining suffix. In the fixed-length block view shown in Figure~\ref{fig:misalign}, a boundary token may also enter or leave the block, so the resulting block can have a Levenshtein distance greater than one from the target. Corruption positions and inserted tokens are seeded independently of model predictions, so both models receive identical corrupted inputs. A block is considered repaired only if the generation obtained after editing exactly matches the target sequence.

\subsection{Pass@$k$ Decoding Configuration}
\label{sec:passk-config}
For the pass@$k$ comparison in Section~\ref{sec:exp-passk-decoding}, both regimes use the same checkpoint, temperature $0.6$, block size $32$, and a maximum generation length of $32{,}768$. We draw 256 samples per problem. Confidence-free decoding uses $\tau=0$ with \Method{} and permits at most 16 post-mask editing rounds; thresholded decoding uses $\tau=0.95$ without editing, matching the previous standard dLLM setup. The number of decoding iterations is dynamic: every iteration decodes at least one masked token, rather than using a fixed iteration count.

\subsection{Benchmark Suite}
\label{app:bench}
We evaluate on $12$ benchmarks spanning seven capability categories, including knowledge, reasoning, math, coding, function calling, instruction following, and Chinese, as summarized in Table~\ref{tab:bench}. All benchmarks are evaluated in a zero-shot setting. We use temperature $0.0$ for the reported pass@1 scores; the separate pass@$k$ comparison uses temperature $0.6$ (Appendix~\ref{sec:passk-config}). We report pass@1 for every benchmark. Answer extraction and scoring follow each benchmark's native format. For math tasks, boxed expressions (\verb|\boxed{}|) are verified either symbolically using MATH-Verify or by an external judge model with xVerify. For multiple-choice tasks, we use regex-matched letter answers. For code tasks, we rely on unit-test execution, including EvalPlus tests and sandboxed runs for LiveCodeBench. For KOR-Bench, IFEval, and BFCL v3, we use their task-specific rule-based scorers.

\begin{table}[h]
\caption{Evaluation benchmarks, grouped by capability category.}
\label{tab:bench}
\begin{center}
\small
\begin{tabular}{@{}l l@{}}
\toprule
Category & Benchmark \\
\midrule
\multirow{2}{*}{Knowledge} & MMLU-Pro \citep{mmlu-pro} \\
                           & GPQA-Diamond \citep{gpqa} \\
\midrule
Reasoning & KOR-Bench \citep{kor-bench} \\
\midrule
\multirow{3}{*}{Math} & GSM-Plus \citep{gsm-plus} \\
                      & OlympiadBench \citep{olympiadbench} \\
                      & AIME 2025 \citep{aime} \\
\midrule
\multirow{3}{*}{Coding} & HumanEval+ \citep{evalplus} \\
                        & MBPP+ \citep{evalplus} \\
                        & LiveCodeBench v6 \citep{livecodebench} \\
\midrule
Function calling & BFCL v3 \citep{bfcl} \\
\midrule
Instruction following & IFEval (prompt-strict) \citep{ifeval} \\
\midrule
Chinese & C-Eval \citep{ceval} \\
\bottomrule
\end{tabular}
\end{center}
\end{table}

\section{Additional Related-Work Comparisons}
\label{app:related-comparison}

\textbf{LLaDA2.1 token editing.}
LLaDA2.1 allows previously generated tokens to be replaced during masked denoising, but each update remains tied to the original block positions \citep{llada21}.
\Method{} is post-trained from the same 16B LLaDA2.1-Mini Base and extends this action space with deletion and insertion.
A deletion removes a committed token, whereas an insertion creates a new mask whose content is resolved by a later mask-to-token update; the resulting state is then projected back to the original block size.

\textbf{ME-DLM.}
ME-DLM is the closest edit-based refinement method: it first produces a complete masked-diffusion draft and subsequently applies replacement, deletion, and insertion to the resulting variable-length sequence \citep{ren2026editbased}.
It adds a next-token prediction head alongside current-token prediction, and its training pipeline comprises three sequential stages.
In contrast, \Method{} represents token prediction, deletion, and insertion in a single per-position output vocabulary and uses one post-training objective along an inference-like denoising trajectory.
Insertion creates a mask rather than directly predicting its content, allowing \Method{} to reuse the base DLM's token predictor and to correct alignment before block generation is complete, without a separate post-draft refinement phase.

\textbf{DreamOn and FlexMDM.}
DreamOn introduces expansion and deletion states for length-adaptive code infilling, with both states predicted at masked positions and the sequence length changing during inference \citep{dreamon}.
FlexMDM instead alternates mask insertion and unmasking over a variable-length state and predicts an additional insertion expectation for each gap \citep{kim2025flexmdm}.
These methods primarily use length-changing transitions to model or control output length.
\Method{} uses them to correct model-induced intermediate states: it can edit already visible tokens, requires no separate insertion-expectation head, and keeps each model call compatible with the fixed-block interface of its base DLM.

\textbf{Edit-based generative processes.}
Edit Flows defines insertion, deletion, and substitution as transition rates of a continuous-time generative process over variable-length sequences \citep{havasi2025edit}.
Unlike this approach, \Method{} does not replace the underlying generative process; it adds discrete correction actions through post-training of an existing masked DLM.
The experimental scales also differ: ME-DLM and the scaled FlexMDM use LLaDA-8B Base, DreamOn uses several 7B DLMs, and Edit Flows primarily reports 1.3B language models trained from scratch, whereas our experiments use the later 16B LLaDA2.1-Mini Base.
Because these backbones and training settings are not matched, we treat their scale as implementation context rather than as evidence of cross-paper performance superiority.

\section{Detailed Results}
\label{app:training_ablation}

\begin{figure}[!htb]
\centering
\includegraphics[width=0.95\textwidth]{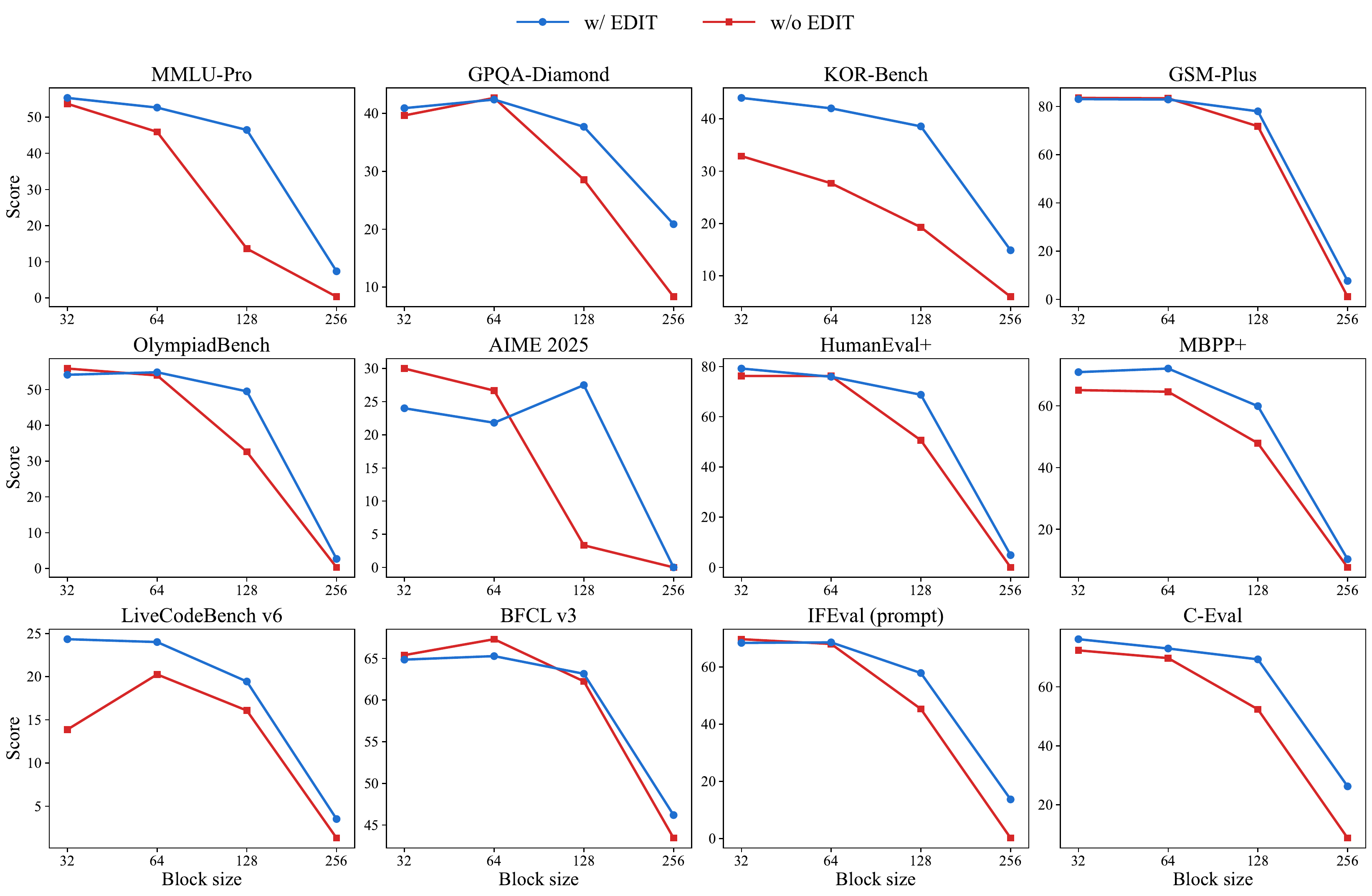}
\caption{Detailed performance across different block sizes for the baseline and \Edit{}.}
\label{fig:block-pb}
\end{figure}

Figure~\ref{fig:block-pb} presents the detailed per-benchmark performance across different block sizes for the 12 benchmarks introduced in Section~\ref{sec:exp-block}. As shown in the figure, \Edit{} outperforms the baseline on the majority of benchmarks, demonstrating that \Method{} can greatly mitigate error accumulation within extended denoising chains by containing local alignment errors before they propagate into block-level corruption.

\begin{figure}[!htb]
  \centering
  \begin{subfigure}[b]{0.45\textwidth}
    \centering
    \includegraphics[width=\linewidth]{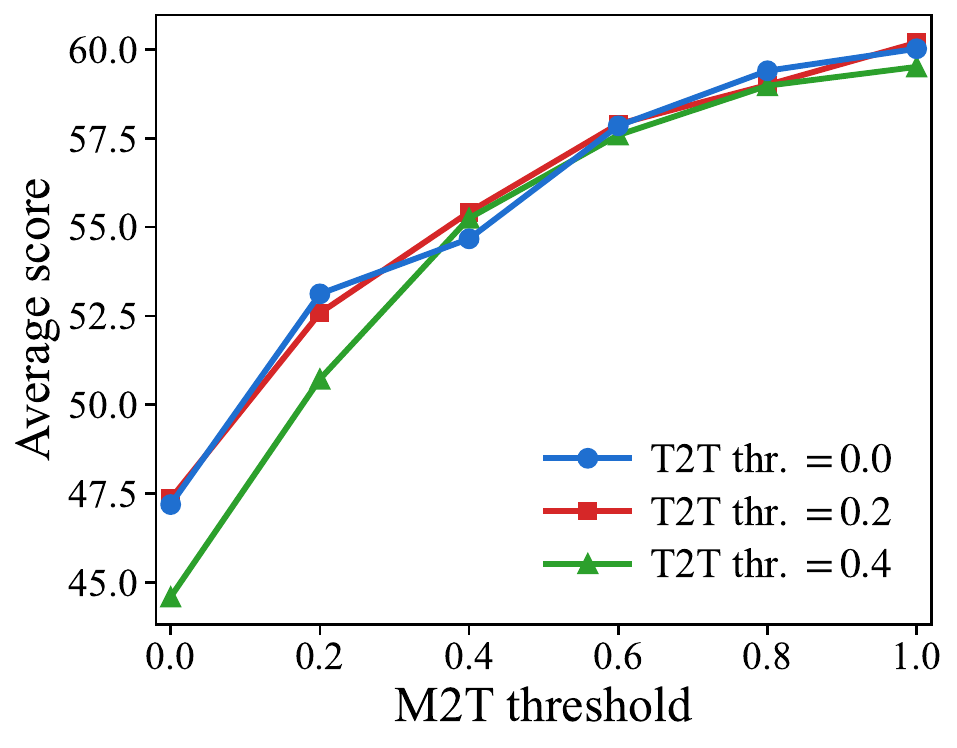}
    \caption{Score across configurations.}
    \label{fig:grid-score}
  \end{subfigure}%
  \hfill
  \begin{subfigure}[b]{0.45\textwidth}
    \centering
    \includegraphics[width=\linewidth]{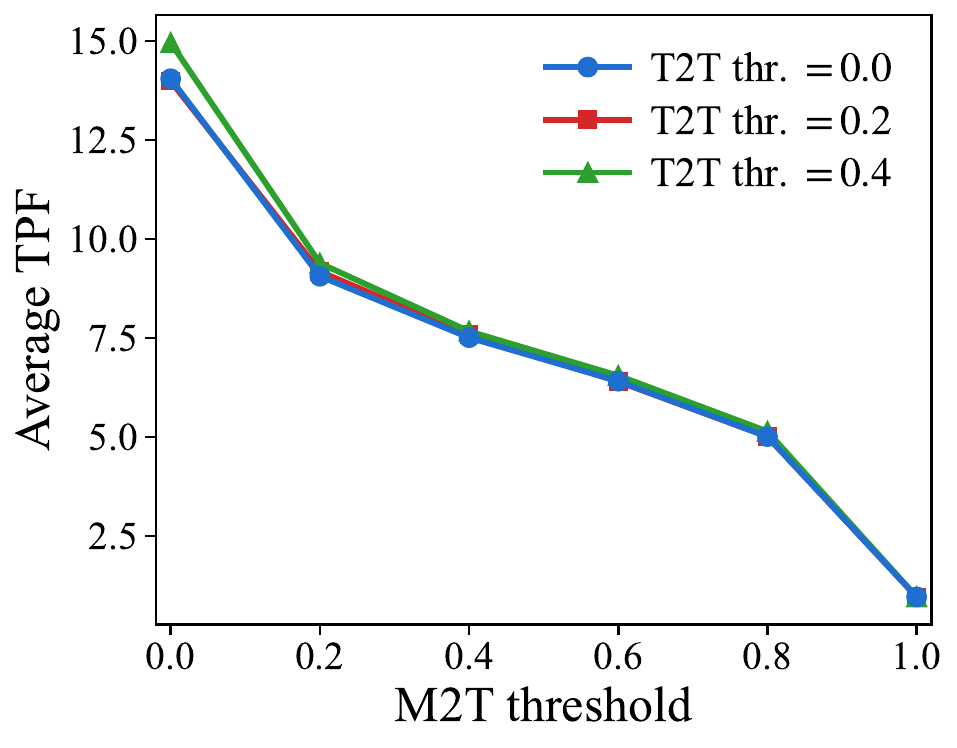}
    \caption{TPF across configurations.}
    \label{fig:grid-tpf}
  \end{subfigure}
  \caption{Aggregate grid-search results with block size fixed at $32$.}
  \label{fig:grid}
\end{figure}

\begin{figure}[!htb]
\centering
\includegraphics[width=0.95\textwidth]{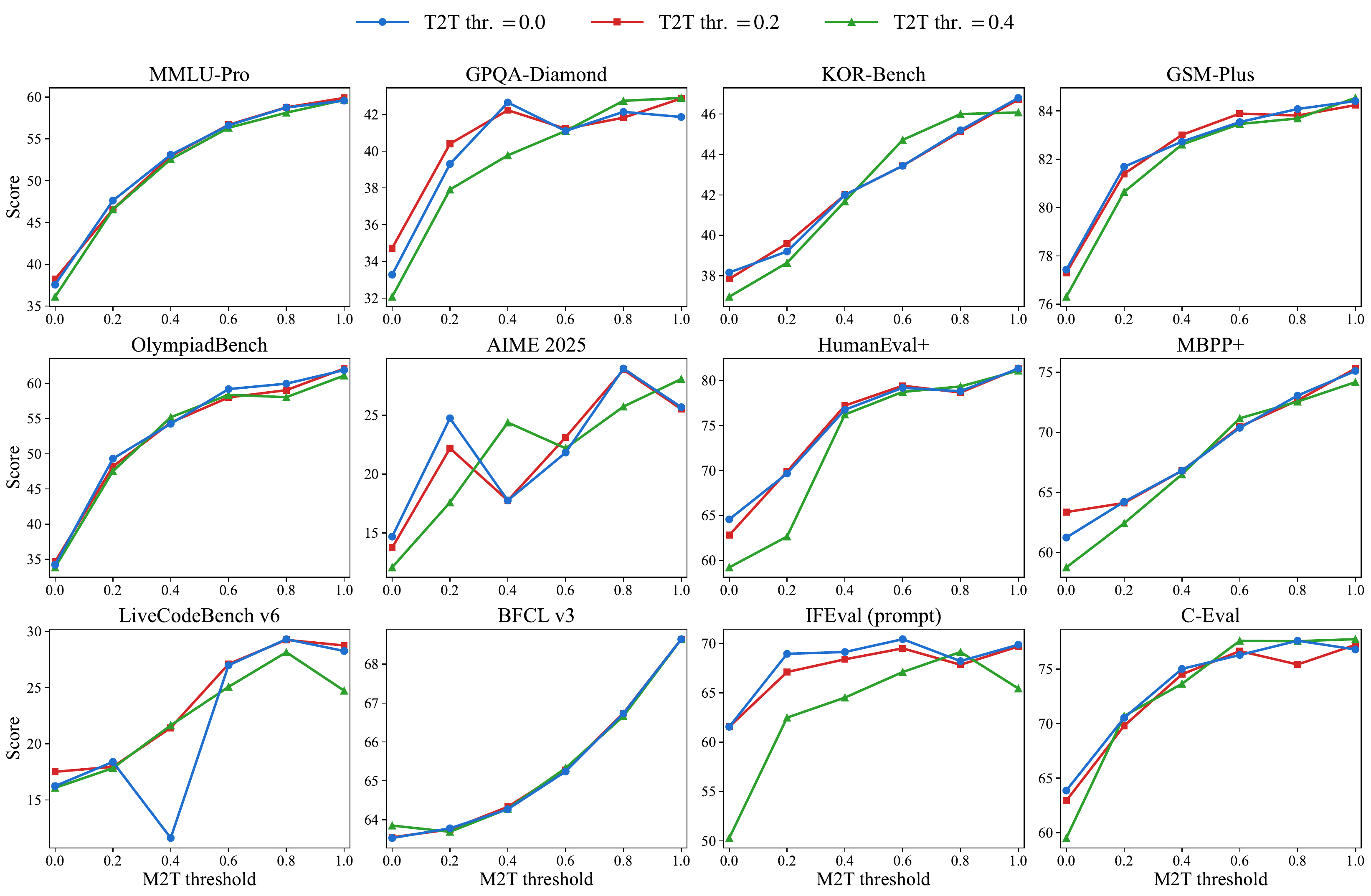}
\caption{Per-benchmark scores across the mask and edit threshold grid at temperature $0.0$.}
\label{fig:grid_per_bc}
\end{figure}

Figure~\ref{fig:grid} reports the aggregate grid-search results used to select the default thresholds in Section~\ref{sec:exp-grid}. Figure~\ref{fig:grid_per_bc} provides the corresponding per-benchmark breakdown.

\section{Method Details and Algorithms}
\label{app:algs}
This section complements Section~\ref{sec:method} with the formal details needed to reproduce each component. For each procedure, we first state its contract and boundary behavior in prose and then give pseudocode that makes the execution order explicit.

\paragraph{Forming the next input block (Algorithm~\ref{alg:block-update}).}
At round $r$, the model predicts one token from $\Vout$ at every position. The procedure uses the predictions at the selected positions $\mathcal{Q}_r$. It retains $s_i^r$ outside $\mathcal{Q}_r$. At a selected position, $\Del$ removes $s_i^r$, $\Ins$ places a fresh $\Mask$ immediately before $s_i^r$, and a token in $\Voc$ replaces $s_i^r$. The temporary sequence may have a length different from $B$. Keeping its first $B$ tokens and padding a shorter sequence on the right with $\Mask$ gives the input $s^{r+1}$ for the next round.

\begin{algorithm}[!htb]
\caption{Form the input block for round $r+1$}
\label{alg:block-update}
\KwIn{input block $s^r$; selected positions $\mathcal{Q}_r\subseteq\{1,\ldots,B\}$; one predicted token in $\Vout$ for each $i\in\mathcal{Q}_r$; block length $B$}
\KwOut{input block $s^{r+1}$ for round $r+1$}
$\mathrm{seq}\leftarrow[\,]$\;
\For{$i=1\ldots B$}{
  \uIf{$i\notin\mathcal{Q}_r$}{append $s_i^r$ to $\mathrm{seq}$\;}
  \uElseIf{the prediction at $i$ is $\Del$}{do not append $s_i^r$\;}
  \uElseIf{the prediction at $i$ is $\Ins$}{append $\Mask$, then $s_i^r$, to $\mathrm{seq}$\;}
  \Else{append the predicted token in $\Voc$ to $\mathrm{seq}$\;}
}
keep the first $B$ tokens of $\mathrm{seq}$; if it has fewer than $B$ tokens, append $\Mask$ tokens on the right until it has length $B$\;
$s^{r+1}\leftarrow\mathrm{seq}$\;
\Return $s^{r+1}$\;
\end{algorithm}

\paragraph{LCS-based edit-target construction (Algorithm~\ref{alg:lcs}).}
Given the input block $s^r$ and target block $y=(y_1,\ldots,y_B)$, the procedure constructs an edit target $e_i^r\in\Vout$ for every position $i$. It first obtains the maximum-cardinality monotone matching $\pi^\star$ defined in Section~\ref{sec:method-lcs}. The matched pairs divide the remaining positions into the ordered gap pairs $(I_k,J_k)$. Matched or paired positions receive their token from $y$, extra positions in $s^r$ receive $\Del$, and a nonfinal gap with extra positions in $y$ assigns one $\Ins$ target to its right matched anchor. Extra target positions in the final gap are handled in a later round.

\begin{algorithm}[!htb]
\caption{Construct edit targets by LCS alignment}
\label{alg:lcs}
\KwIn{input block $s^r$ and target block $y=(y_1,\ldots,y_B)$}
\KwOut{edit targets $e_i^r\in\Vout$ for $i=1,\ldots,B$}

$\pi^\star\leftarrow$ a maximum-cardinality monotone matching of positions in $s^r$ and $y$ where $s_i^r=y_j\in\Voc$ for every matched pair $(i,j)$\;
If several matchings have the same cardinality, break ties by the indices in $s^r$ first and then by the indices in $y$.\;

$K\leftarrow|\pi^\star|$\;
Write $\pi^\star=\{(i_k,j_k)\}_{k=1}^{K}$ and let $(I_k,J_k)$ for $k=0,\ldots,K$ be the ordered gap pairs.\;

\For{$k=1,\ldots,K$}{
  $e_{i_k}^r\leftarrow y_{j_k}$\;
}

\For{$k=0,\ldots,K$}{
  $\ell_k\leftarrow\min\{|I_k|,|J_k|\}$\;

  \For{$p=1,\ldots,\ell_k$}{
    $e_{I_k[p]}^r\leftarrow y_{J_k[p]}$\;
  }

  \For{$p=\ell_k+1,\ldots,|I_k|$}{
    $e_{I_k[p]}^r\leftarrow\Del$\;
  }

  \If{$|J_k|>\ell_k \land k<K$}{
    $e_{i_{k+1}}^r\leftarrow\Ins$\;
  }
}

\Return $(e_i^r)_{i=1}^{B}$\;
\end{algorithm}

\paragraph{Edit-augmented inference (Algorithm~\ref{alg:infer}).}
At round $r$, the decoder updates the positions in $\mathcal{Q}_r=\mathcal{C}_r(\tau)\cup\mathcal{E}_r(\nu)$, using the sets defined in Section~\ref{sec:method-infer}. It samples an output token at each selected position and forms the next block as described in Algorithm~\ref{alg:block-update}. Positions outside $\mathcal{Q}_r$ retain their current tokens.

After all masks are resolved, the decoder permits at most $K_{\max}=16$ additional rounds and stops early when $\mathcal{E}_r(\nu)=\emptyset$. The decoder records previously visited block states in $\mathcal{H}$ and resamples a selected output token if an update would repeat a state. Algorithm~\ref{alg:infer} gives the complete procedure.

\begin{algorithm}[!htb]
\caption{Decode a block with edit-augmented transitions}
\label{alg:infer}
\KwIn{context $c$, initial state $s^0=\Mask^B$, mask threshold $\tau$, edit threshold $\nu$, temperature $T$, maximum post-mask rounds $K_{\max}=16$}
\KwOut{decoded block state $s^r$}
$\mathcal{H}\leftarrow\{s^0\}$ \quad $r\leftarrow0$ \quad $k\leftarrow0$\;
\While{true}{
  $p_r\leftarrow p_\theta(\cdot\mid c,s^r)$\;
  $\mathcal{M}_r\leftarrow\{i\in\{1,\ldots,B\}:s_i^r=\Mask\}$ \quad $\mathcal{P}_r\leftarrow\{i\in\{1,\ldots,B\}:s_i^r\in\Voc\}$\;
  \For{$i=1,\ldots,B$}{$\hat{v}_i^r\leftarrow\argmax_{v\in\Vout}p_\theta(v\mid c,s^r,i)$\;}
  \For{$i\in\mathcal{P}_r$}{$\eta_i^r\leftarrow p_\theta(\hat{v}_i^r\mid c,s^r,i)$\;}
  $\mathcal{C}_r(\tau)\leftarrow$ mask-update set defined in Section~\ref{sec:method-infer}\;
  $\mathcal{E}_r(\nu)\leftarrow\{i\in\mathcal{P}_r:\hat{v}_i^r\neq s_i^r\ \land\ \eta_i^r\geq\nu\}$\;
  \If{$\mathcal{M}_r=\emptyset\ \land\ \big(\mathcal{E}_r(\nu)=\emptyset\ \lor\ k=K_{\max}\big)$}{\textbf{break}\;}
  $\mathcal{Q}_r\leftarrow\mathcal{C}_r(\tau)\cup\mathcal{E}_r(\nu)$\;
  \For{$i\in\mathcal{Q}_r$}{sample an output token $\tilde{v}_i^r$ from $p_\theta(\cdot\mid c,s^r,i)$ at temperature $T$\;}
  $\tilde{s}\leftarrow$ apply the sampled outputs to $s^r$ as in Algorithm~\ref{alg:block-update}\;
  \While{$\tilde{s}\in\mathcal{H}$}{
    choose $j$ uniformly from $\mathcal{Q}_r$. Set the probability of $\tilde{v}_j^r$ to zero and resample $\tilde{v}_j^r$\;
    $\tilde{s}\leftarrow$ apply the sampled outputs to $s^r$ as in Algorithm~\ref{alg:block-update}\;
  }
  $s^{r+1}\leftarrow\tilde{s}$ \quad $\mathcal{H}\leftarrow\mathcal{H}\cup\{s^{r+1}\}$\;
  \If{$\mathcal{M}_r=\varnothing$}{$k\leftarrow k+1$\;}
  $r\leftarrow r+1$\;
}
\Return $s^r$\;
\end{algorithm}

\paragraph{Inference-aligned multi-stage training (Algorithm~\ref{alg:train}).}
This loop implements the compressed training trajectory described in Section~\ref{sec:method-multistage}. Starting from $s^0=\Mask^B$, each stage selects masks that satisfy the confidence threshold and supplements them with the highest-confidence remaining masks to meet $m_r=\min\{|\mathcal{M}_r|,\lceil B/R\rceil\}$. It derives LCS labels for the selected masks and visible positions, then compiles the resulting actions into the next state.

Selected masks use mixed teacher forcing with ratio $\rho$, while visible positions advance with their edit labels. The final editing-only stage supervises visible tokens produced by the last mask-update stage, and the returned objective averages the $R+1$ normalized stage losses. Figure~\ref{fig:multistage} illustrates the trajectory, and Algorithm~\ref{alg:train} specifies the stage loop and the final editing-only update.

\begin{algorithm}[!htb]
\caption{Inference-aligned multi-stage training}
\label{alg:train}
\KwIn{context $c$, clean target $y$, block size $B$, stages $R$, threshold $\tau$, teacher-forcing ratio $\rho$}
$s^0\leftarrow\Mask^B$ \quad $\mathcal{P}_0\leftarrow\varnothing$ \quad $\mathcal{L}\leftarrow0$\;
\For{$r=0\ldots R-1$}{
  $p_r\leftarrow p_\theta(\cdot\mid c,s^r)$\;
  $\mathcal{C}_r(\tau)\leftarrow$ confidence-thresholded masked positions from $p_r$\;
  $m_r\leftarrow\min\{|\mathcal{M}_r|,\lceil B/R\rceil\}$ \quad $\mathcal{U}_r\leftarrow\mathcal{C}_r(\tau)$\;
  \If{$|\mathcal{U}_r|<m_r$}{
    $\mathcal{U}_r\leftarrow\mathcal{U}_r\cup\topop_{m_r-|\mathcal{U}_r|}\!\big(\mathcal{M}_r\setminus\mathcal{U}_r\big)$\;
  }
  $\mathcal{Q}_r\leftarrow\mathcal{U}_r\cup\mathcal{P}_r$ \quad $e^r\leftarrow\textsc{BuildEditLabels}(s^r,y)$\;
  $\mathcal{L}\mathrel{+}=\mathcal{L}_r$ on $\mathcal{Q}_r$ via Eq.~\eqref{eq:stage_r_loss}\;
  form $\tilde a^r_{\mathcal{Q}_r}$ by mixed teacher forcing on $\mathcal{U}_r$ (ratio $\rho$) and edit labels $e_i^r$ on $\mathcal{P}_r$\;
  $s^{r+1}\leftarrow\textsc{Compile}(s^r,\tilde a^r_{\mathcal{Q}_r},\mathcal{Q}_r,B)$ \quad $\mathcal{P}_{r+1}\leftarrow\{i:s_i^{r+1}\in\Voc\}$\;
}
$\mathcal{U}_R\leftarrow\varnothing$ \quad $e^R\leftarrow\textsc{BuildEditLabels}(s^R,y)$ \quad $p_R\leftarrow p_\theta(\cdot\mid c,s^R)$\;
$\mathcal{L}\mathrel{+}=\mathcal{L}_R$ on $\mathcal{P}_R$ via Eq.~\eqref{eq:stage_r_loss}\;
\Return $\mathcal{L}/(R{+}1)$\;
\end{algorithm}

\clearpage
\section{Case Study: Edit Trajectories at \texorpdfstring{$\tau=0$}{tau=0}}
\label{app:case}

\definecolor{editred}{RGB}{198,40,40}
\definecolor{editblue}{RGB}{30,80,200}
\definecolor{editgreen}{RGB}{20,130,60}
\definecolor{editorange}{RGB}{200,110,0}
\definecolor{prefixgray}{RGB}{70,70,70}
\newcommand{\ttok}[1]{\texttt{#1}}                            
\newcommand{\tdel}[1]{\textcolor{editred}{\texttt{#1}}}       
\newcommand{\tnew}[1]{\textcolor{editgreen}{\textbf{\texttt{#1}}}} 
\newcommand{\tpre}[1]{\textcolor{prefixgray}{\texttt{#1}}}    
\newcommand{\nlt}{\textcolor{prefixgray}{$\hookleftarrow$}}   
\newcommand{\elc}{\textcolor{prefixgray}{$\cdots$}}           
\newcommand{\spDel}{{\setlength{\fboxsep}{1pt}\colorbox{red!15}{\textcolor{editred}{\footnotesize\textsf{[Del]}}}}}
\newcommand{\spIns}{{\setlength{\fboxsep}{1pt}\colorbox{orange!22}{\textcolor{editorange}{\footnotesize\textsf{[Ins]}}}}}
\newcommand{\spMask}{{\setlength{\fboxsep}{1pt}\colorbox{blue!15}{\textcolor{editblue}{\footnotesize\textsf{[M]}}}}}
\newcommand{\spc}{\textcolor{prefixgray}{$\sqcup$}}          
\newcommand{\sdrop}[1]{\textcolor{editred}{\texttt{#1}}}     
\newcommand{\edittrace}[1]{\par\endgroup\nopagebreak\smallskip\noindent\makebox[\linewidth]{\renewcommand{\arraystretch}{1.3}#1}\par\smallskip}
\newcommand{\prefixline}{\par\nopagebreak\begingroup\raggedright\noindent\textbf{Prefix (fixed):}\;}

We trace six single-block decoding runs drawn from a panel of GSM-Plus problems (each a paraphrase of the same robe task) under the aggressive $\tau=0$ regime, where the model commits every position at each round. Each run applies the same per-token corruption to both \Method{} and a \emph{baseline} (an identical model trained without the edit actions), and we group the runs into three \emph{successful repairs} where \Method{} converges exactly to the target sequence while the baseline fails, and three \emph{failures} where \Method{} does not. Each example states the fixed \emph{prefix} (everything before the editable region) and then traces, over a window of token positions, the row labeled \emph{Target} followed by the \emph{input} and \emph{output} of each step. We show as much of the editable block as fits and elide the rest with \elc{}. Notation: \spDel{} deletes its position, \spIns{} opens a mask before its position, \spMask{} is an empty mask, \nlt{} is a newline, \elc{} marks elided positions; the \tdel{injected corruption} is red and a \tnew{restored} token is green. Because $\Del$/$\Ins$ shift the suffix, only the columns before the first edit are strictly position-aligned across rows.

\subsection{Successful repairs}

\textbf{Example~1} (insertion) is a successful repair at edit distance $0$: a spurious \tdel{rumah} is inserted after \ttok{blue}, which shifts the suffix and drops the final token \ttok{white}; the model removes it with \spDel{} and refills the reopened tail \spMask{} with \tnew{white}.
\prefixline
\tpre{Question: A robe takes 3 bolts of blue fiber and half that much white fiber. How many bolts in total does it take? ... within \textbackslash boxed\{\}. Let's break it down step by step:}\,\nlt\,\nlt\,\tpre{1. The robe requires **3 bolts}
\edittrace{\setlength{\tabcolsep}{3.5pt}
\begin{tabular}{r ccccccccccc}
\emph{position} & 17 & 18 & 19 & 20 & 21 & 22 & 23 & \elc & 29 & 30 & 31 \\
\midrule
Target & \ttok{of} & \ttok{blue} & \ttok{fiber} & \ttok{**.} & \nlt & \nlt & \ttok{2} & \elc & \ttok{as}   & \ttok{much} & \ttok{white} \\
\midrule
Step 0 in     & \ttok{of} & \ttok{blue} & \tdel{rumah} & \ttok{fiber} & \ttok{**.} & \nlt & \nlt & \elc & \ttok{half} & \ttok{as}   & \ttok{much} \\
Step 0 out    & \ttok{of} & \ttok{blue} & \spDel       & \ttok{fiber} & \ttok{**.} & \nlt & \nlt & \elc & \ttok{half} & \ttok{as}   & \ttok{much} \\
\midrule
Step 1 in     & \ttok{of} & \ttok{blue} & \ttok{fiber} & \ttok{**.}  & \nlt & \nlt & \ttok{2} & \elc & \ttok{as}   & \ttok{much} & \spMask \\
Step 1 out    & \ttok{of} & \ttok{blue} & \ttok{fiber} & \ttok{**.}  & \nlt & \nlt & \ttok{2} & \elc & \ttok{as}   & \ttok{much} & \tnew{white} \\
\end{tabular}}

\textbf{Example~2} (deletion) is the harder case where a \emph{deleted} token must be actively re-introduced. The opening \ttok{**} before \ttok{blue} is deleted, so \ttok{blue} slides into the bold marker's slot. \Method{} first rewrites the shifted \ttok{blue} back into \ttok{**} (step~1), then emits \spIns{} to open a slot (step~2) and refills it with \tnew{blue} (step~3); the run converges in five rounds (the leading whitespace-only round is omitted) at edit distance~$0$, while the baseline cannot reopen a slot and settles at edit distance~$2$ with \ttok{fiber} dropped.
\prefixline
\tpre{Question: A robe takes 200 bolts of blue fiber ... within \textbackslash boxed\{\}. Let's break this down step by step:}\,\nlt\,\nlt
\edittrace{\setlength{\tabcolsep}{3.5pt}
\begin{tabular}{r cccccccccc}
\emph{position} & 11 & 12 & 13 & 14 & 15 & 16 & 17 & 18 & 19 & 20 \\
\midrule
Target & \ttok{1} & \ttok{.} & \ttok{The} & \ttok{number} & \ttok{of} & \ttok{**} & \ttok{blue} & \ttok{fiber} & \ttok{**} & \ttok{bolts} \\
\midrule
Step 1 in   & \ttok{1} & \ttok{.} & \ttok{The} & \ttok{number} & \ttok{of} & \tdel{blue} & \ttok{fiber} & \ttok{**} & \ttok{bolts} & \ttok{needed} \\
Step 1 out  & \ttok{1} & \ttok{.} & \ttok{The} & \ttok{number} & \ttok{of} & \ttok{**} & \ttok{fiber} & \ttok{**} & \ttok{bolts} & \ttok{needed} \\
\midrule
Step 2 in   & \ttok{1} & \ttok{.} & \ttok{The} & \ttok{number} & \ttok{of} & \ttok{**} & \ttok{fiber} & \ttok{**} & \ttok{bolts} & \ttok{needed} \\
Step 2 out  & \ttok{1} & \ttok{.} & \ttok{The} & \ttok{number} & \ttok{of} & \ttok{**} & \spIns & \ttok{fiber} & \ttok{**} & \ttok{bolts} \\
\midrule
Step 3 in   & \ttok{1} & \ttok{.} & \ttok{The} & \ttok{number} & \ttok{of} & \ttok{**} & \spMask & \ttok{fiber} & \ttok{**} & \ttok{bolts} \\
Step 3 out  & \ttok{1} & \ttok{.} & \ttok{The} & \ttok{number} & \ttok{of} & \ttok{**} & \tnew{blue} & \ttok{fiber} & \ttok{**} & \ttok{bolts} \\
\end{tabular}}

\Needspace{16\baselineskip}
\textbf{Example~3} (deletion) follows the same \spIns{}-then-refill pattern more directly: the token \ttok{It} is deleted, pulling \ttok{also} into its slot, and \Method{} emits \spIns{} before \ttok{also} and refills the mask with \tnew{It}, reaching edit distance~$0$ in three steps where the baseline fails at edit distance~$2$.
\prefixline
\tpre{Question: A robe takes 2 bolts of blue fiber and 0.5 times ... within \textbackslash boxed\{\}. Let's break this down step by step:}\,\nlt\,\nlt\,\tpre{- The robe requires **2 bolts}
\edittrace{\setlength{\tabcolsep}{3pt}
\begin{tabular}{r cccccccccc}
\emph{position} & 18 & 19 & 20 & 21 & 22 & 23 & 24 & 25 & 26 & 27 \\
\midrule
Target & \ttok{of} & \ttok{blue} & \ttok{fiber} & \ttok{**.} & \nlt & \ttok{-} & \ttok{It}   & \ttok{also} & \ttok{requires} & \ttok{**} \\
\midrule
Step 0 in  & \ttok{of} & \ttok{blue} & \ttok{fiber} & \ttok{**.} & \nlt & \ttok{-} & \tdel{also} & \ttok{requires} & \ttok{**} & \ttok{0} \\
Step 0 out & \ttok{of} & \ttok{blue} & \ttok{fiber} & \ttok{**.} & \nlt & \ttok{-} & \spIns       & \ttok{also} & \ttok{requires} & \ttok{**} \\
\midrule
Step 1 in  & \ttok{of} & \ttok{blue} & \ttok{fiber} & \ttok{**.} & \nlt & \ttok{-} & \spMask      & \ttok{also} & \ttok{requires} & \ttok{**} \\
Step 1 out & \ttok{of} & \ttok{blue} & \ttok{fiber} & \ttok{**.} & \nlt & \ttok{-} & \tnew{It}    & \ttok{also} & \ttok{requires} & \ttok{**} \\
\end{tabular}}

\subsection{Failures}
The runs that do not converge cluster into three distinct failure modes. The first is \emph{inaction on a locally fluent corruption}.

\textbf{Example~4} (deletion, edit distance $1$): the token \ttok{'s} is deleted, pulling \ttok{break} into position~31; because \ttok{Let break} is not flagged as anomalous by the local context, the model keeps \ttok{break} and converges in one step instead of rewriting it to \tnew{'s} or predicting \spIns{} to re-insert it.
\prefixline
\tpre{Question: A robe takes 3 bolts of blue fiber and half that much white fiber. How many bolts in total does it take? Please reason step by step, and put your final answer within \textbackslash boxed\{\} <|role\_end|>}
\edittrace{\setlength{\tabcolsep}{7pt}
\begin{tabular}{r ccccccc}
\emph{position} & 25 & 26 & 27 & 28 & 29 & 30 & 31 \\
\midrule
Target & \ttok{<role>} & \ttok{ASS} & \ttok{IST} & \ttok{ANT} & \ttok{</role>} & \ttok{Let} & \ttok{'s} \\
\midrule
Step 0 in     & \ttok{<role>} & \ttok{ASS} & \ttok{IST} & \ttok{ANT} & \ttok{</role>} & \ttok{Let} & \tdel{break} \\
Step 0 out    & \ttok{<role>} & \ttok{ASS} & \ttok{IST} & \ttok{ANT} & \ttok{</role>} & \ttok{Let} & \ttok{break} \\
\end{tabular}}

The second mode is \emph{commitment to the corruption}: rather than deleting the junk, the model treats it as a valid prefix and extends it.

\textbf{Example~5} (insertion): an inserted \tdel{Ric} before \ttok{fiber} is completed into the word \tdel{Richet} by a token-to-token rewrite that overwrites \ttok{fiber}, cementing the error at edit distance~$3$.
\prefixline
\tpre{Question: A robe takes 2 bolts of blue fiber and 0.5 times ... within \textbackslash boxed\{\}. Let's break this down step by step:}\,\nlt\,\nlt\,\tpre{- The robe requires **}
\edittrace{\setlength{\tabcolsep}{5pt}
\begin{tabular}{r cccccccc}
\emph{position} & 16 & 17 & 18 & 19 & 20 & 21 & 22 & 23 \\
\midrule
Target & \ttok{2} & \ttok{bolts} & \ttok{of} & \ttok{blue} & \ttok{fiber} & \ttok{**.} & \nlt & \ttok{-} \\
\midrule
Step 0 in  & \ttok{2} & \ttok{bolts} & \ttok{of} & \ttok{blue} & \tdel{Ric} & \ttok{fiber} & \ttok{**.} & \nlt \\
Step 0 out & \ttok{2} & \ttok{bolts} & \ttok{of} & \ttok{blue} & \tdel{Ric} & \tdel{het}   & \ttok{**.} & \nlt \\
\end{tabular}}

The third mode is \emph{over-deletion}.

\Needspace{16\baselineskip}
\textbf{Example~6} (insertion): an inserted \tdel{Flint} is removed, but \Method{} fires a second \spDel{} on the adjacent valid \ttok{of}, then refills with a spurious \ttok{**}, so the genuine token \ttok{blue} is never restored.
\prefixline
\tpre{Question: A robe takes 2 bolts of blue fiber ... Each fiber costs \$12 ... within \textbackslash boxed\{\}. Let's break this down step by step.}\,\nlt\,\nlt\,\tpre{1. **Blue fiber used**:}\,\nlt\,\tpre{The robe takes **}
\edittrace{\setlength{\tabcolsep}{5pt}
\begin{tabular}{r cccccccc}
\emph{position} & 21 & 22 & 23 & 24 & 25 & 26 & 27 & 28 \\
\midrule
Target & \ttok{The} & \ttok{robe} & \ttok{takes} & \ttok{**} & \ttok{2} & \ttok{bolts} & \ttok{of}    & \ttok{blue} \\
\midrule
Step 0 in  & \ttok{The} & \ttok{robe} & \ttok{takes} & \ttok{**} & \ttok{2} & \ttok{bolts} & \tdel{Flint} & \ttok{of} \\
Step 0 out & \ttok{The} & \ttok{robe} & \ttok{takes} & \ttok{**} & \ttok{2} & \ttok{bolts} & \spDel       & \spDel \\
\midrule
Step 1 in  & \ttok{The} & \ttok{robe} & \ttok{takes} & \ttok{**} & \ttok{2} & \ttok{bolts} & \spMask      & \spMask \\
Step 1 out & \ttok{The} & \ttok{robe} & \ttok{takes} & \ttok{**} & \ttok{2} & \ttok{bolts} & \tdel{**}    & \ttok{of} \\
\end{tabular}}

\subsection{Synthesis}
Taken together, the six runs show that \Method{} has genuinely learned to \emph{use} all three edit actions and, unlike the baseline, can repair both spurious insertions (with $\Del$) and outright deletions (with $\Ins$), often reaching the target sequence exactly where the action-free baseline cannot. The failures are not random: they concentrate where the corrupted text is locally fluent so no anomaly signal fires (the \ttok{break} case), or where the model commits to or over-extends the corruption rather than excising it (the \ttok{Richet} and \ttok{Flint} cases). The common thread is that the model edits along a \emph{locally} plausible path rather than the globally shortest one, which is exactly what the local LCS labels supervise. This motivates stronger path-level supervision, such as offline edit-path construction that scores whole recovery trajectories, beyond the per-position LCS labels used here.

\end{document}